\pdfoutput=1

\documentclass[11pt]{article}

\usepackage{ACL2023}

\usepackage{times}
\usepackage{latexsym}

\usepackage[T1]{fontenc}

\usepackage[utf8]{inputenc}

\usepackage{microtype}

\usepackage{inconsolata}

\usepackage{multirow}
\usepackage{graphicx}
\usepackage{CJKutf8}
\usepackage{subcaption}
\usepackage{marvosym}

\title{Topic-driven Distant Supervision Framework for Macro-level \\ Discourse Parsing }

\author{Feng Jiang$^1$, Longwang He$^2$, Peifeng Li$^2$, Qiaoming Zhu$^2$, Haizhou Li$^1$ \\
  $^1$ School of Data Science, The Chinese University of Hong Kong, Shenzhen, China \\
  $^2$ School of Computer Science and Technology, Soochow University, Suzhou, China  \\
  \texttt{\{jeffreyjiang, haizhouli\}@cuhk.edu.cn} \\
  \texttt{lwhe@stu.suda.edu.cn}\\
  \texttt{\{pfli,qmzhu\}@suda.edu.cn}\\
  }

\begin{document}
\maketitle
\begin{abstract}
Discourse parsing, the task of analyzing the internal rhetorical structure of texts, is a challenging problem in natural language processing. Despite the recent advances in neural models, the lack of large-scale, high-quality corpora for training remains a major obstacle. Recent studies have attempted to overcome this limitation by using distant supervision, which utilizes results from other NLP tasks (e.g., sentiment polarity, attention matrix, and segmentation probability) to parse discourse trees. However, these methods do not take into account the differences between in-domain and out-of-domain tasks, resulting in lower performance and inability to leverage the high-quality in-domain data for further improvement. To address these issues, we propose a distant supervision framework that leverages the relations between topic structure and rhetorical structure. Specifically, we propose two distantly supervised methods, based on transfer learning and the teacher-student model, that narrow the gap between in-domain and out-of-domain tasks through label mapping and oracle annotation. Experimental results on the MCDTB and RST-DT datasets show that our methods achieve the best performance in both distant-supervised and supervised scenarios.
\end{abstract}

\section{Introduction}\label{sec:introduction}

Every discourse unit (e.g., a clause, sentence, or paragraph) is semantically closely connected in a coherent document. Discourse parsing is discovering the internal structure of the whole document formed by these discourse units, which is beneficial to many NLP applications, such as automatic summarization~\cite{cohan2018scientific}, reading comprehension~\cite{mihaylov-frank-2019-discourse}, and machine translation~\cite{tan2022towards}.

As one of the most influential theories of discourse analysis, Rhetorical Structure Theory (RST)~\cite{mann1987rhetorical} represents a document as a hierarchical Discourse Tree (DT). In general, a discourse tree can be split into micro and macro levels~\cite{van1983strategies}. Compared with the micro-level structure focusing on the relationship between clauses and sentences, the macro-level structure focuses on that between paragraphs or chapters to understand the full document at a higher level, which is more important for downstream tasks~\cite{kobayashi-etal-2021-improving}.

Since the existing manual annotated corpora~\cite{carlson2003building,subba-di-eugenio-2009-effective,jiang-etal-2018-MCDTB} only contain several hundred documents, their small size constraints further improve the performance of the supervised deep neural network~\cite{zhang-etal-2021-adversarial, DBLP:conf/aaai/JiangFCLZ021, yu-etal-2022-rst}, especially at the macro level. Due to the large annotation granularity and the complex annotation object, it is time-consuming and expensive to annotate a large-scale, high-quality discourse corpus manually. 

Therefore, mainstream research attempts to use other NLP tasks to distantly supervise parsing discourse trees with fewer domain-dependent data. Earlier work used sentiment analysis~\cite{huber-carenini-2019-predicting,huber-carenini-2020-MEGA} and summarization~\cite{xiao-etal-2021-predicting} tasks to build discourse trees with distant supervision. \citet{huber2022predicting} proposed a topic-driven distantly supervised method, achieving better performance. They trained a topic segmentation model and converted the results into discourse trees from top to down according to the order of topic segmentation probability.

\begin{figure}[htbp]
\centerline{\includegraphics[width=8cm]{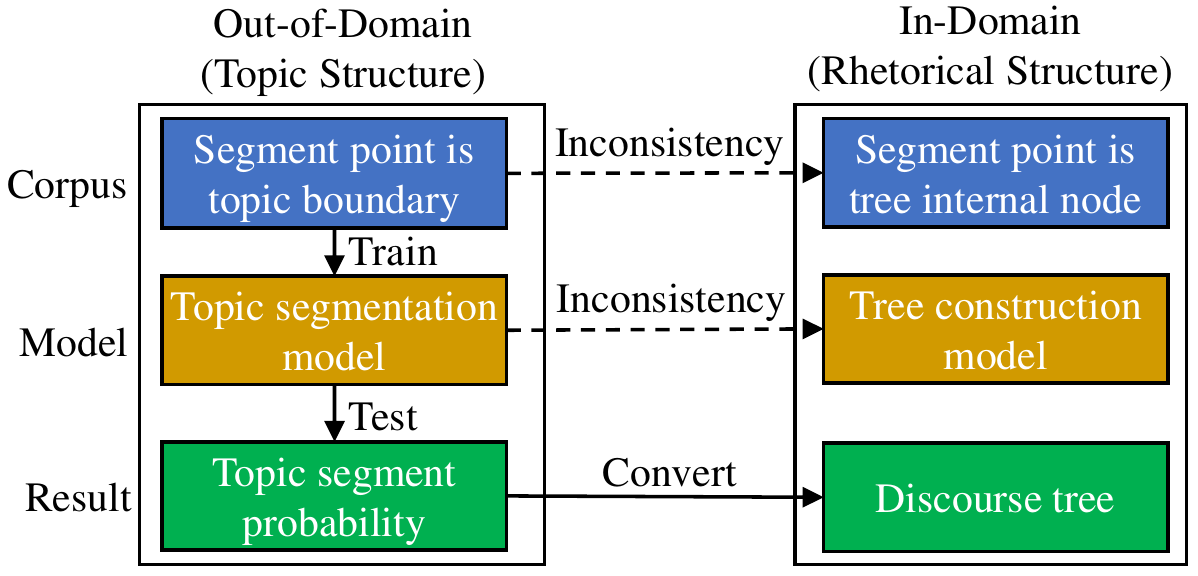}}
\caption{The topic-driven distant supervision by result converting. The models and corpora are inconsistent between in- and out-of-domain tasks. Appendix~\ref{sec:appendixd} shows the details process of the result converting.}
\label{huber}
\end{figure}

However, they do not take into account the inconsistency of the model and corpus between in- and out-of-domain, as shown in Figure~\ref{huber}. It brings two challenges to break the bottleneck of discourse parsing due to the different learning objectives and annotation forms between in- and out-of-domain tasks. The first challenge is that there is still a large gap between the performance of distant supervised and supervised methods. The second challenge is that the distant supervised models cannot leverage in-domain high-quality annotations for further improvement. 

To solve the above issues, we propose a topic-driven distant supervision framework, as shown in Figure~\ref{fig_framework}. Based on the previous work~\cite{huber2022predicting}, we further propose two novel distant supervised methods in the framework to narrow the distance between in-domain and out-of-domain tasks, which utilizes two internal relations between topic structure and rhetorical structure~\cite{DBLP:conf/aaai/JiangFCLZ021}: 1) Locally, if two adjacent discourse units have a rhetorical relation, they are likely to belong to the same topic. 2) Globally, the topic structure is the skeleton of the rhetorical structure tree, and each topic contains a discourse sub-tree. 

Unlike the result converting method ~\cite{huber2022predicting}, we first propose a transfer learning method following the first relation. It uses label mapping to keep the consistency of the model. Furthermore, we propose a teacher-student model following the second relation. It uses the teacher model to construct the silver rhetorical corpus by oracle annotation to ensure the consistency of the corpus. Moreover, since our proposed methods introduce models in the in-domain task, we can use high-quality in-domain data for retraining the model to make further improvements. Experimental results on the Chinese MCDTB and English RST-DT corpora show that our method achieves the best performance in both distant and supervised scenarios, demonstrating the effectiveness of our proposed framework. 

\section{Related Work}

\subsection{Topic Segmentation}
Topic segmentation~\cite{hearst-1997-text} aims to mine topic maintenance and shift in text, and is generally formalized as determining whether each part is the boundary of a topic given a text sequence. With the large-scale topic of corpora construction (e.g., WIKI727K~\cite{koshorek-etal-2018-text}), supervised methods based on deep learning, especially based on the pre-trained models, are more popular. 

One research line used sequential labeling to predict topic boundaries with different double-layer neural models, such as CNN+LSTM~\cite{badjatiya2018attention}, LSTM+LSTM~\cite{koshorek-etal-2018-text} and transformer+transformer~\cite{glava2020twolevel}. Thanks to the pre-trained language model,  \citet{lukasik-etal-2020-text} proposed three models using BERT as sentence encoder (i.e., cross BERT, BERT+bi-LSTM and hierarchy BERT) to improve the performance of topic segmentation, while \citet{DBLP:conf/aaai/JiangFCLZ021} provided a TM-BERT as a local model with slide window to predict topic boundaries. 

Another research line on topic segmentation used point network to predict topic boundaries. For example, ~ \citet{li-2018-segbot} proposed a segbot model, which encodes text through the gated recurrent unit (GRU) module and used a pointer network to obtain topic segmentation points. \citet{xing-etal-2020-improving} combined sequential labeling of topic segmentation with local coherence modeling to achieve better performance.

\begin{figure*}[htbp]
\centerline{\includegraphics[width=16cm]{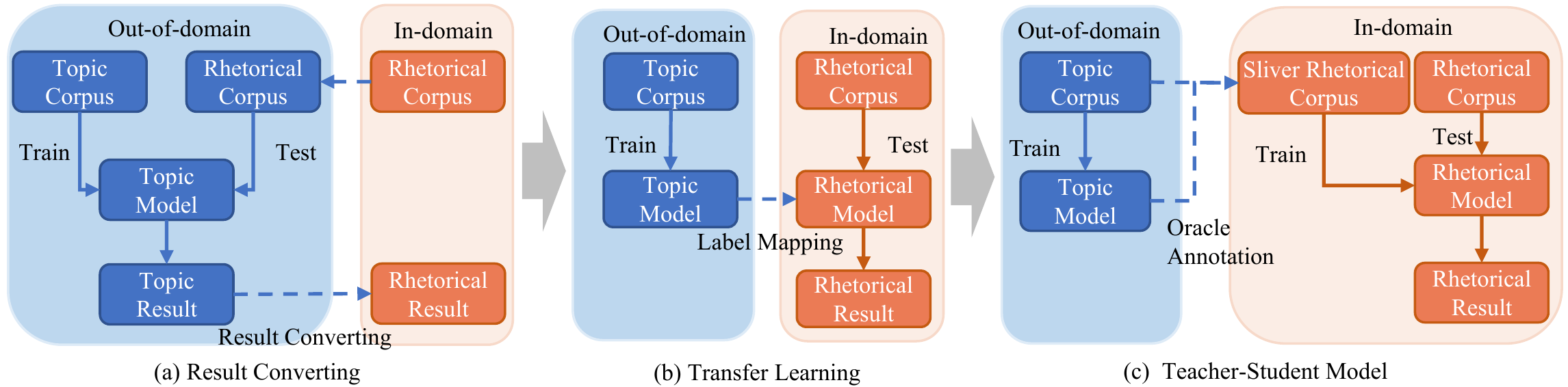}}
\caption{The Topic-driven distant supervision framework for discourse parsing. It contains three methods according to the distance between in- and out-of-domain tasks: (a) Result Converting, (b) Transfer Learning and (c) Teacher-Student Model.}
\label{fig_framework}
\end{figure*}

\subsection{Distantly Supervised Discourse Parser}

Compared with the flat topic structure, the hierarchy rhetorical structure is more complex. Due to the lack of large-scale manually annotated corpora, existing studies attempt to construct discourse trees that are distantly supervised by other tasks.

\citet{huber-carenini-2019-predicting,huber-carenini-2020-MEGA} used distant supervision to generate discourse trees from sentiment analysis. They leveraged the relation between the sentiment polarity of children and parents through multiple-case learning to construct discourse rhetorical structure trees. \citet{xiao-etal-2021-predicting} constructed discourse trees that are distantly supervised by summarization. They obtained the association between each Elementary Discourse Unit (EDU) by attention matrix in the transformer-based summary model and parsed the discourse tree through the CYK and CLE algorithms. \citet{huber2022predicting} built discourse trees using distant supervision based on topic segmentation. They greedily constructed a discourse tree from top to down by order of topic segmentation probabilities.

\section{Topic-driven Distant Supervision Framework for Macro-level Discourse Parsing}

We propose a topic-driven distant supervision framework for macro-level discourse parsing based on the internal relations between topic structure and rhetorical structure, as shown in Figure~\ref{fig_framework}. It includes three methods: result converting, transfer learning, and teacher-student model.

The result converting method (Figure~\ref{fig_framework} (a)) was proposed by ~\citet{huber2022predicting}. Although this is a successful attempt at topic-driven distant supervision for discourse parsing, it does not consider the two internal relations between topic structure and rhetorical structure.

Following the first internal relation, we propose a distantly supervised method based on transfer learning, as shown in Figure~\ref{fig_framework} (b). To ensure the consistency of the learning objective, it unifies the rhetorical structure and topic structure prediction of the two adjacent discourse units into discourse coherence prediction. Thus, the topic segmentation model is transformed into a rhetorical structure tree construction model, which can be used to construct intra-domain discourse trees via the shift-reduce algorithm.

Furthermore, we propose a distantly supervised method based on the teacher-student model following the second internal relation, as shown in Figure~\ref{fig_framework} (c). The teacher model (topic segmentation model) first uses the oracle annotation to construct a silver rhetorical structure corpus by the golden topic structure. Then, the student model (rhetorical structure tree construction model) can learn from it to build the discourse tree, ensuring the consistency of the annotated information between in- and out-of-domain tasks. 

\subsection{Distantly Supervised Discourse Parser on Transfer Learning}

\begin{figure}[htbp]
\centerline{\includegraphics[width=7.5cm]{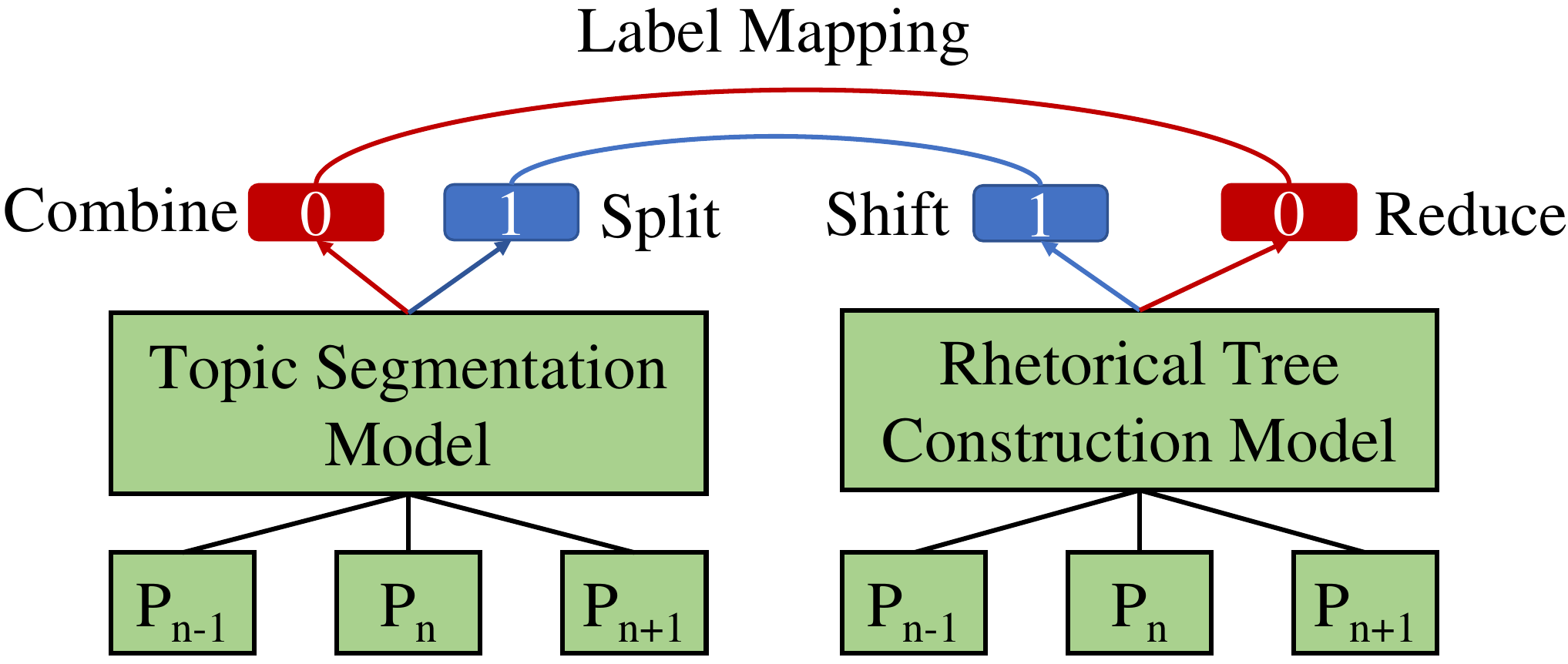}}
\caption{The architecture of distantly supervised discourse parser based on transfer learning.}
\label{fig_transfer}
\end{figure}

To address the inconsistency between in- and out-of-domain learning objectives, we propose a distantly supervised discourse parser based on transfer learning. Instead of converting results, it converts models from out-of-domain to in-domain using the discourse coherence consistency of labels between the topic segmentation model and the rhetorical tree construction model, as shown in Figure~\ref{fig_transfer}.

\begin{figure*}[htbp]
\centerline{\includegraphics[width=15cm]{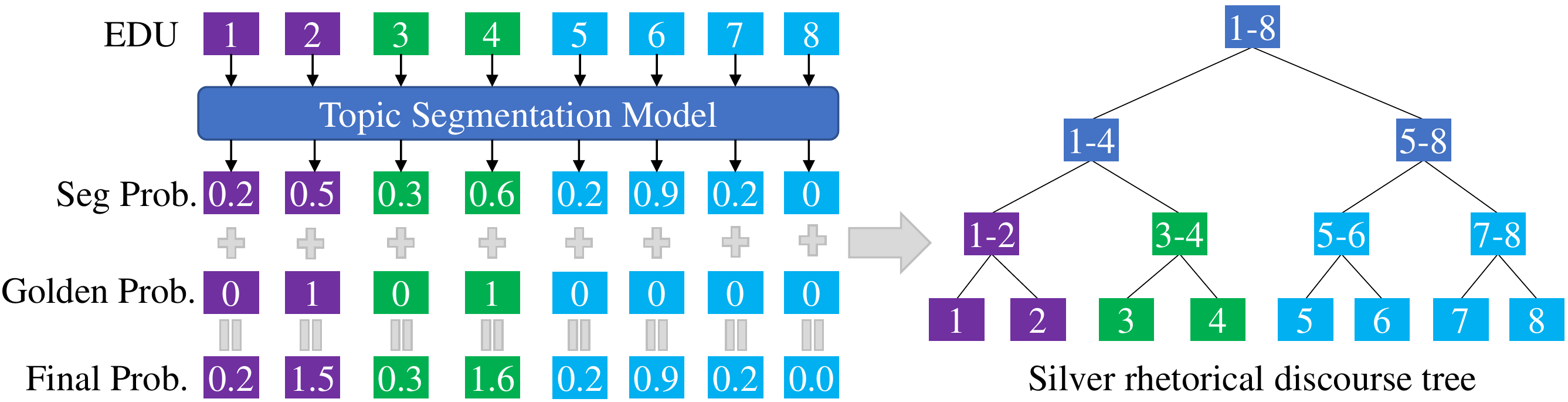}}
\caption{The oracle annotation in teacher model. The EDUs in the same topic have the same color. Seg Prob. indicates the segmentation probability of each EDU predicted by the topic segmentation model, Golden Prob. indicates the topic boundary probability of each EDU (last one ignored), and Final Prob. indicates the split probability in the oracle annotation to build the discourse tree from top to down.}
\label{fig_teacher}
\end{figure*}

In the out-of-domain task, we adopt the sequential labeling model~\cite{DBLP:conf/aaai/JiangFCLZ021}, which uses a local TM-BERT model to segment topics through sliding windows. For each discourse unit, the model needs to predict whether it is the boundary of the topic according to the context. The predicted results are labeled as \textit{combine} or \textit{split}.

In the in-domain task, we adopt the transition-based parser~\cite{DBLP:conf/aaai/JiangFCLZ021}, which views the discourse tree construction into a sequence of actions containing the \textit{shift} and \textit{reduce}. According to the first relation, we map the labels of the topic segmentation model to the transition-based parser through the coherence between two adjacent discourse units to maintain the consistency of the learning objective, as shown in Table~\ref{tab:labelmapping}. 

\begin{table}[htbp]
\begin{center}
\resizebox{\linewidth}{!}{
\begin{tabular}{llll}
\hline
\textbf{Learning Objective}  & \textbf{Topic Model}  & \textbf{Rhetorical Model} & \textbf{Label} \\ \hline
 coherent & combine    & reduce                         & 0               \\
 incoherent & split   & shift                         & 1               \\ \hline
\end{tabular}
}
\caption{The label mapping in transfer learning method.}\label{tab:labelmapping}
\end{center}
\vspace{-0.4cm}
\end{table}

\begin{figure}[htbp]
\centerline{\includegraphics[width=7.5cm]{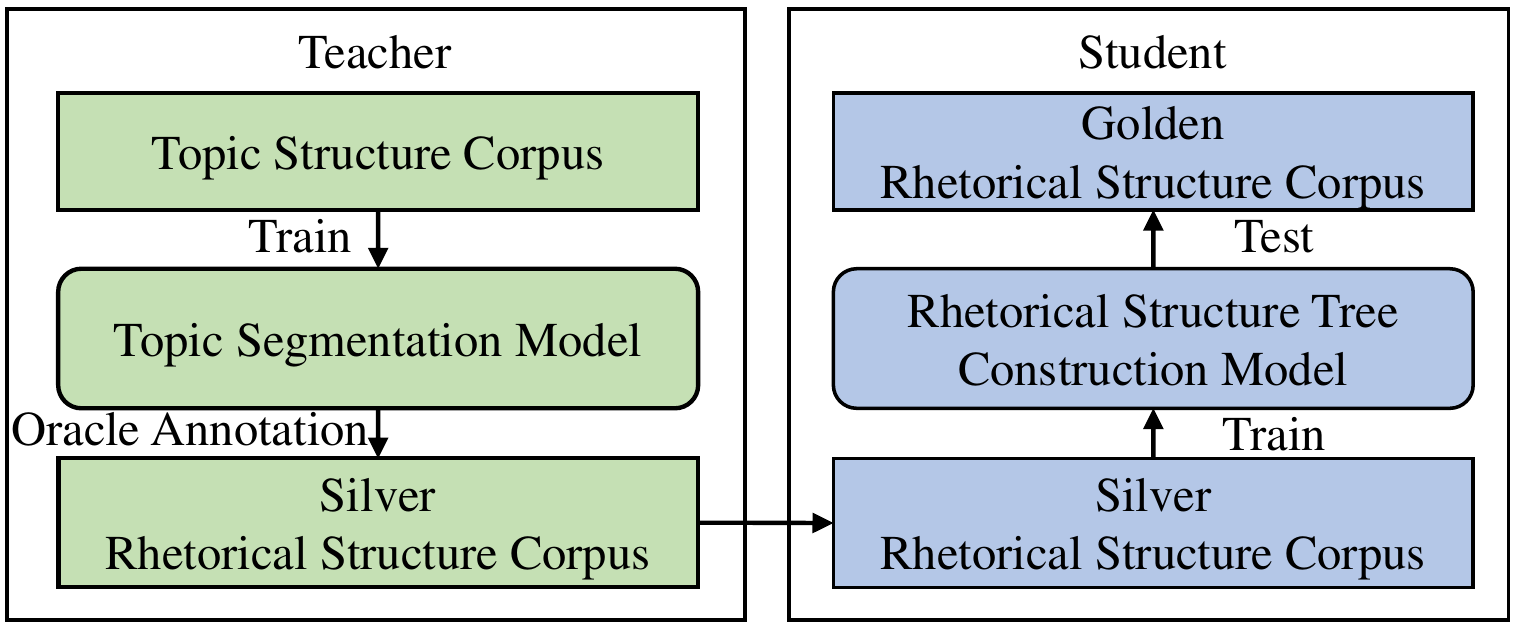}}
\caption{The architecture of the teacher-student model.}
\label{fig_teacher_student}
\end{figure}

\subsection{Distantly Supervised Discourse Parser on Teacher-Student Model}

To address the inconsistency of the annotation form, we further propose a distantly supervised discourse parser based on the teacher-student model according to the second relation, as shown in Figure~\ref{fig_teacher_student}.

In the out-of-domain task, we first use a topic segmentation model as the teacher model trained on the topic structure corpus. We then propose the oracle annotation to build rhetorical trees with golden topic boundaries for obtaining a silver rhetorical structure corpus. 

In the in-domain task, we bring this silver rhetorical structure corpus and propose a bi-directional pointer network BLINK as the student model to learn rhetorical structures from it. Therefore, we unify the in-domain and out-of-domain annotation forms through the oracle-annotated silver rhetorical structure corpus and build the exclusive student and teacher model in the in- and out-of-domain tasks, narrowing the distance between them.

\subsubsection{Teacher Model}\label{sec:teacher model}

The result converting method~\cite{huber2022predicting} offers the possibility of using the topic segmentation model to construct rhetorical structure trees but its improvement is limited, while ~\citet{DBLP:conf/aaai/JiangFCLZ021} used golden topic structures to assist discourse parsing and achieved a higher accuracy (about 83\%). Inspired by the above two methods, we propose an oracle annotation method to build a silver rhetorical structure corpus with higher quality based on the golden topic structure, as shown in Figure~\ref{fig_teacher}.

First, we use a topic segmentation model\footnote{Here, we use TM-BERT ~\cite{DBLP:conf/aaai/JiangFCLZ021}. Although we have tried other models (e.g., BERT+Bi LSTM and Hier. BERT), TM-BERT achieves the highest performance.} to predict the probability of each topic segment point, following previous work~\cite{huber2022predicting}. Then, different from it, we use the golden topic boundary as the constraint condition according to the second relation, greedily building the silver discourse rhetorical structure tree from top to down by the final probability (Final Prob.). It can ensure that the constructed discourse rhetorical tree is better. For example, EDU6 in Figure~\ref{fig_teacher} should be the first split point with the highest segmentation probability (Seg Porb.) to generate a wrong tree if we do not consider the golden topic boundary, while the silver rhetorical discourse tree constructed by oracle annotation is better due to using the golden probability (Golden Prob.).

We conduct a ten-fold cross-validation of the out-of-domain topic structure corpus to get the silver discourse trees by the teacher model with the oracle annotation. In each fold, we use nine parts as training sets and the rest one part as the test set. More details about the silver rhetorical corpus are shown in Appendix~\ref{sec:appendixa}.

\subsubsection{Student Model}
\begin{figure}[htbp]
\centerline{\includegraphics[width=7.5cm]{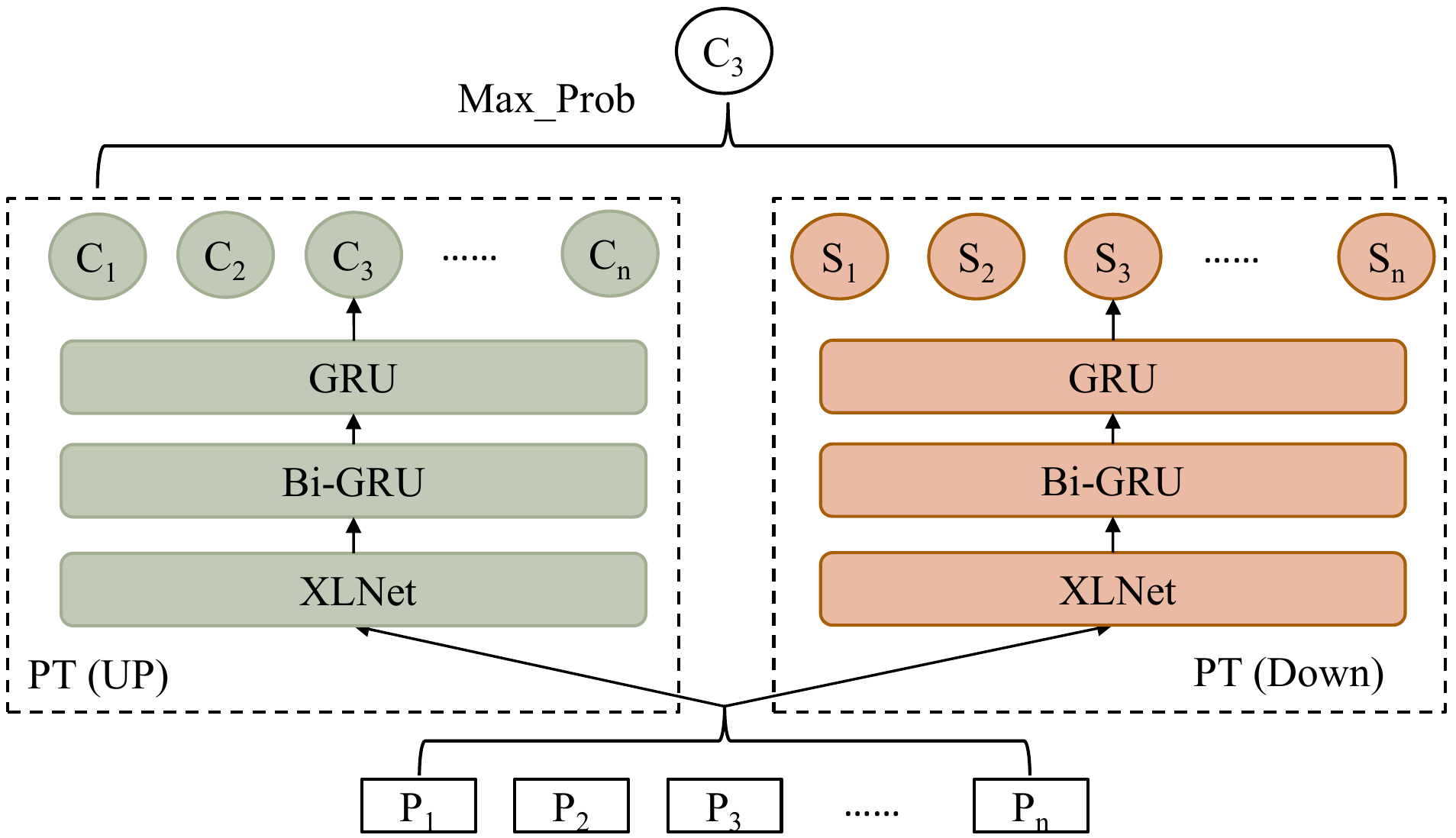}}
\caption{The architecture of student model (BLINK).}
\label{fig_BLINK}
\end{figure}

Inspired by previous work~\cite{kobayashi-etal-2019-split}, we propose a bi-directional pointer network (BLINK) as the student model. The model consists of two popular pointer networks: a top-down split network (PT (Down)) and a bottom-up merge network (PT (Up)). When building a discourse rhetorical tree, the final operation of each step is determined by the maximum probability of the prediction of two networks, as shown in Figure~\ref{fig_BLINK}.

PT (Down) and PT (Up) have the same architecture. In the encoder, we first use the pre-trained model XLNet to encode all paragraphs of the document. Then, we use XLNet to obtain the vector representation of each word in the input $W= \{w_1, w_2,..., w_m \} $, where $m$ represents the number of words input in the document. After that, we use the Bi-GRU module to encode $W$ to obtain the overall semantic representation of the document $E= \{e_1, e_2,..., e_m \} $, as shown in Eq. \ref{eq:bigru}. Then, at each step $t$, we obtain the vector of each \textit{<SEP>} token as the representation of paragraphs $ P= \{p_ {<t, 1>}, p_ {<t,2>},...,p_ {<t, n>} \}$, where $n$ is the number of paragraphs included in a document. 

\begin{equation}\label{eq:bigru}
E, h_f=f_{Bi-GRU}(W, h_0)
\end{equation}

At the decoding step $t$, we feed the vector of the last paragraph ($p_ {<t, l>} $) and the hidden layer vector $ h_{t-1} $ of the decoder in the previous time step into the decoder (GRU) to obtain the decoding representation ($d_t$) of the current discourse units sequence, as shown in Eq. \ref{eq:gru}.

\begin{equation}\label{eq:gru}
d_t,h_t=f_{GRU}(p_{<t,l>},h_{t-1})
\end{equation}

Finally, we calculate the attention score ($score$) between $d_t$ and each paragraph through dot product ($\delta $) to obtain the probability distribution of the split point or merge point at the current time step (t), as shown in Eqs. \ref{score}, \ref{C} and \ref{S}. 
\vspace{-0.2cm}
\begin{equation}\label{score}
score^m_{<t,i>}=\delta(d_t^m,p_{<t,i>}^m) \quad m \in {c,s}
\end{equation}
\begin{equation}\label{C}
C_{p}=argmax_p(Softmax(score^c_t))
\end{equation}
\begin{equation}\label{S}
S_{p}=argmax_p(Softmax(score^s_t))
\end{equation}
where $C_p=\{c_1, c_2,..., c_n\}$ represents the probability distribution of each paragraph as the combination point, $S_p=\{s_1, s_2,..., s_n $\} indicates the probability distribution of each paragraph as the split point. We select the highest probability value from C$_p$ and S$_p$ as the final action. For example, as shown in Figure \ref{fig_BLINK}, at this step, the BLINK model finally selects the maximum probability value ($C_{3}$), which means that paragraph 3 ($P_3$) and paragraph 4 ($P_4$) should be combined.

\section{Experimentation}
\subsection{Datasets and Matrics}
We first verify the effectiveness of the proposed method on Chinese MCDTB, which contains 720 documents annotated with macro discourse rhetorical structure. Following previous work~\cite{DBLP:conf/aaai/JiangFCLZ021}, we split the dataset into a train set (80\%) and a testing set (20\%). 

To obtain the out-of-domain topic structure corpus, we collected 14393 Xinhua news documents from the Gigaword corpus\footnote{https://catalog.ldc.upenn.edu/LDC2009T2}. Each document has subheadings as topic boundaries. Then we use oracle annotation on it to create the silver rhetorical structure corpus (MCDTB\_dist), as mentioned in Section~\ref{sec:teacher model}. It is worth noting that our methods only use the train data from the MCDTB\_dist in the distantly supervised scenario, while our methods use both the train data of the in-domain (MCDTB) and out-of-domain corpus (MCDTB\_dist) in the supervised scenario. 

The evaluation method consists of previous work~\cite{morey-etal-2017-much,DBLP:conf/aaai/JiangFCLZ021}, which is a more rigorous approach by evaluating the span accuracy. The details of the experimental setup are shown in Appendix~\ref{sec:appendixb}.

\subsection{Baselines}

\textbf{Distantly Supervised Method.}

\textbf{Dist (Paragraph)} method and \textbf{Dist (Topic)} method~\cite{huber2022predicting}. Since there are no existing distantly supervised methods in Chinese, we choose these two English distant supervised methods based on result converting. The former uses the paragraph boundaries in the in-domain corpus as the learning objective, while the latter uses the topic boundaries in the out-of-domain corpus as the learning objective. We reproduce them with TM-BERT in Chinese for a fair comparison.

\textbf{Supervised Model.}

\textbf{BERT} method~\cite{devlin-etal-2019-bert}. Bert is a popular model in various NLP tasks, and we take it as the simple classification local model in the parser.

\textbf{PDParser (w/o TS)} model and \textbf{PDParser (w/ auto TS)} model~\cite{DBLP:conf/aaai/JiangFCLZ021}. They are two SOTA models in Chinese discourse parsing, which are based on a triple semantic matching BERT model (TM-BERT). Their difference is that \textbf{PDParser (w/ auto TS)} model has the predicted topic boundaries to help build discourse trees while \textbf{PDParser (w/o TS)} did not have that. 

\subsection{Results}

The experimental results are shown in Table~\ref{tab:dist_mcdtb}. In distant supervised methods, our transfer learning model and the teacher-student model achieve 56.41\% and 61.51\%, which are 1.08\% and 6.18\% higher than the best baseline Dist (Topic). Moreover, the gradual increase of performance (55.33/56.41/61.51) among the results converting, transfer learning, and the teacher-student model also proves our proposed methods can narrow the distance between in- and out-of-domain tasks.

\begin{table}[htbp]
\centering
\resizebox{\linewidth}{!}{
\begin{tabular}{lll}
\hline
\textbf{Scenario}      & \textbf{Method}        & \textbf{Span}    \\ \hline
Distantly Supervised & Dist (Paragraph)       & 50.23          \\
Distantly Supervised & Dist (Topic)           & 55.33          \\
Distantly Supervised & Transfer Learning (ours)            & 56.41          \\
Distantly Supervised & Teacher-Student (ours)      & \underline{61.51} \\ \hline
Supervised           & BERT                  & 57.19          \\
Supervised           & PDParser (w/o TS)     & 63.06          \\
Supervised           & PDParser (w/ auto TS) & 66.31          \\
Supervised           & Transfer Learning (ours)            & 66.15          \\
Supervised           & Teacher-Student (ours)      & \underline{\textbf{68.01}} \\\hline
\end{tabular}
}
\caption{The performance on MCDTB.}\label{tab:dist_mcdtb}
\end{table}

In addition, in supervised methods, our transfer learning model and teacher-student model, which re-train in- and out-of-domain training data, have achieved 66.15\% and 68.01\% performance, respectively. They are 3.09\% and 4.95\% higher than the baseline model PDParser (w/o TS), and the teacher-student model even gets 1.7\% higher than the SOTA baseline PDParser (w/ auto TS). It shows that the proposed method can fully exploit both in- and out-of-domain annotations, breaking the bottleneck of discourse parsing.

\section{Analysis}
\subsection{Ablation Study}
We perform an ablation study of our proposed teacher-student model to demonstrate its effectiveness, as shown in Table~\ref{tab:ablation}. The first three lines are teacher-student models with different student models. The fourth and fifth lines are two models (Transfer Learning and Dist(Topic)) both using TM-BERT as the local model and topic boundary as the learning objective. The last two lines are two supervised baselines trained only on in-domain data. 

\begin{table}[htbp]
\begin{center}
\resizebox{\linewidth}{!}{
\begin{tabular}{llll}
\hline
\textbf{Method}  & \textbf{Model}  & \textbf{Dist.} & \textbf{Sup.} \\ \hline
Teacher-Student & BLINK    & 61.51                         & 68.01               \\
Teacher-Student & PT (up)   & 57.96                         & 66.00               \\
Teacher-Student & PT (down) & 58.89                         & 64.14               \\ \hline
Transfer Learning       & TM-BERT + Topic   & 56.41                         & -                   \\ 
Dist (Topic)     & TM-BERT + Topic   & 55.33                         & -                   \\ \hline
-     & BLINK            & -                             & 63.37               \\
PDParser (w/o TS)      & TM-BERT          & -                             & 63.06               \\ \hline
\end{tabular}
}
\caption{The ablation study of our proposed methods.}\label{tab:ablation}
\end{center}
\vspace{-0.2cm}
\end{table}

First, the significant improvement between our BLINK and the two single-direction models in both distantly supervised and supervised scenarios shows the effectiveness of our proposed model as a student model. Since it can select the most reliable result from different direction models to build a better discourse structure tree, BLINK achieves the best performance (61.51/68.01) both in distantly supervised and supervised scenarios.

Moreover, whether in supervised or distantly supervised scenarios, even the single-direction models (PT(up) and PT(down)) can outperform the baseline models, proving our proposed framework's effectiveness.

Lastly, since our teacher-student model can leverage both the in- and out-of-domain data to do a re-train, it gets further improvement and is 4.64\% higher than BLINK that only trains on in-domain data. It demonstrates that our created out-of-domain silver rhetorical structure corpus is beneficial for re-training the supervised method.

\subsection{The Effect of Re-training in Different Layers of the Discourse Tree}

\begin{figure}[htbp]
\centerline{\includegraphics[width=7.5cm]{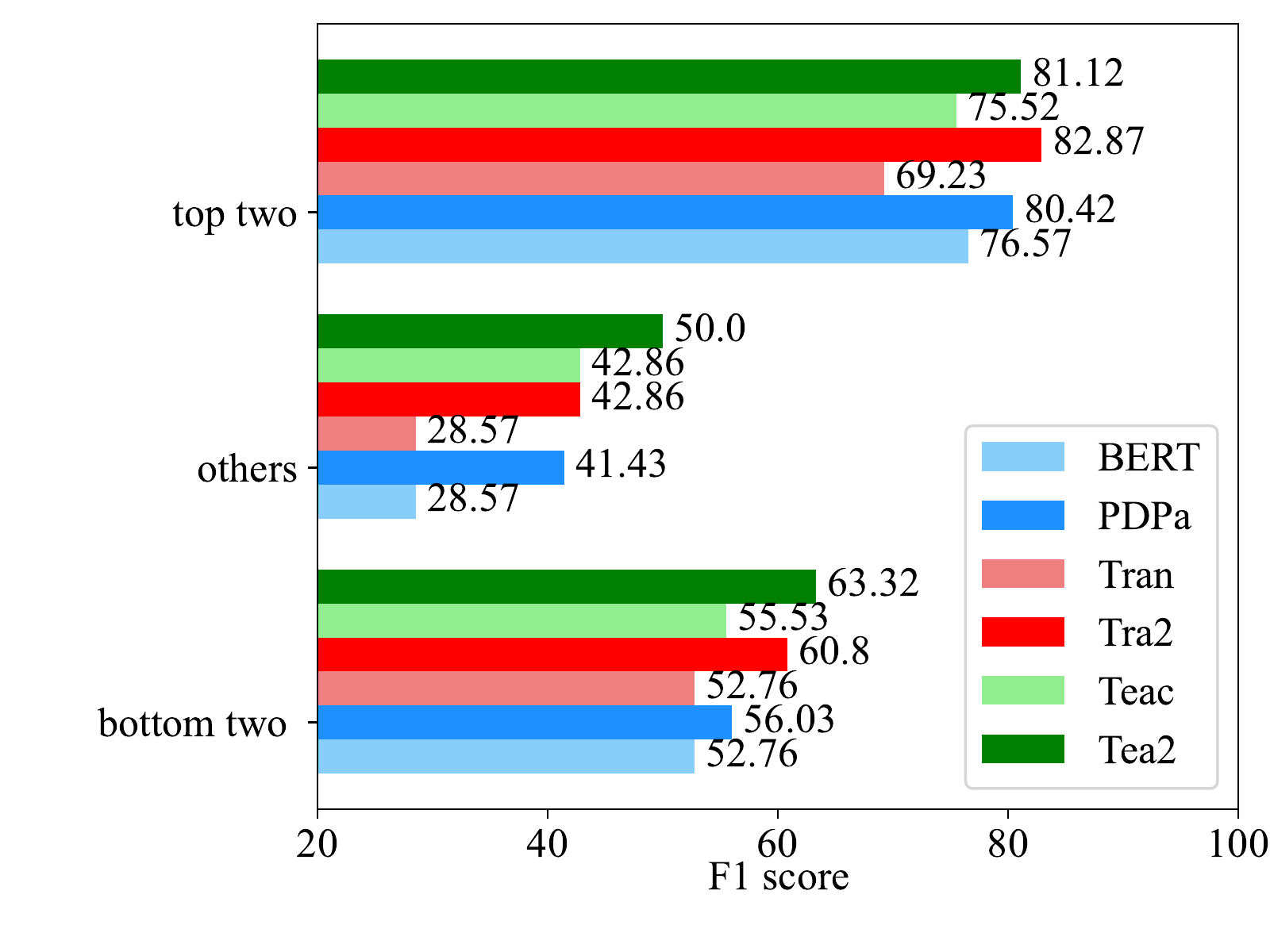}}
\caption{The performance of various models in different layers.}
\label{fig_statich}
\end{figure}

Since our model unifies the learning objective and annotation form, and further improves by re-training in supervised scenarios using the silver corpus, we further analyze the effect of re-training in different layers of the discourse tree, as shown in Figure~\ref {fig_statich}. 

First of all, the distant supervised models based on Transfer Learning (Tran) and Teacher-Student (Teac) are comparable to the two supervised learning models (BERT and PDParser (w/o TS) (PDPa)) on the bottom two layers and the middle layers, while it is slightly weaker than the supervised learning model on the top two layers. For example, compared with BERT and PDPa, the Teac model decreases by 1.05\% and 4.90\%, respectively. The main reason is that topic structure only guarantees the correctness of middle-level boundaries of discourse rhetorical structure, but not that of boundaries higher than topics.

Secondly, the two-stage re-training model (Tra2 and Tea2) can fully use in-domain high-quality annotation information to make up for this defect, achieving better performance. The re-training model (Tra2) based on transfer learning makes further improvement in the middle layer and the top two layers, with an increase of 13.64\% and 14.29\%, respectively, while the teacher-student model (Tea2) makes further improvement in the middle layer and the bottom two layers, with an increase of 7.14\% and 7.79\% respectively.

\subsection{The Effect of Out-of-Domain Corpus in Supervision}

\begin{figure}[htbp]
\centerline{\includegraphics[width=7.3cm]{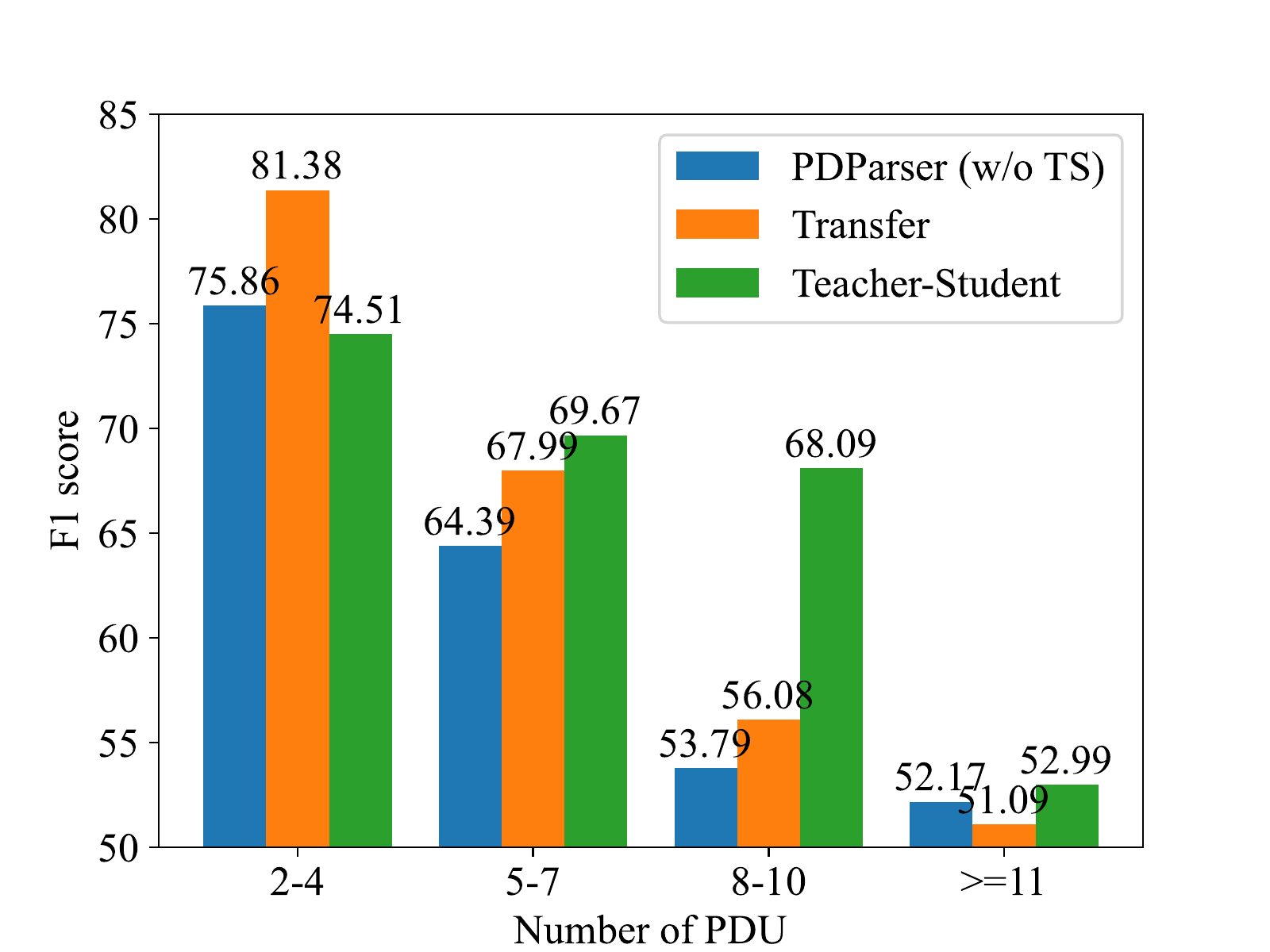}}
\caption{The performance on different length documents without in-domain topic information.}
\label{fig_static_teacher}
\end{figure}

Since our proposed methods also gain a significant improvement with the out-of-domain corpus under the supervised scenario, we further analyze the effect of out-of-domain corpus on constructing structure trees of different length documents, as shown in Figure \ref{fig_static_teacher}.

The transfer learning model achieves 81.38\%, 67.99\%, and 56.08\% in documents with 2-10 paragraphs. Compared with the supervised baseline model (PDParser (w/o TS)), it has improved significantly in shorter documents with 2-4 paragraphs, reaching 5.52\%. We believe that the transfer learning model learns the topic structure better through label mapping due to the relatively simple topic structure of short documents, outperforming the baseline model on these documents.

In addition, the teacher-student model reaches 69.67\%, 68.09\%, and 52.99\%, respectively, in documents with more than five paragraphs and increases significantly in 5-10 paragraph documents by 5.28\% and 14.30\% than the baseline model. Moreover, our teacher-student model is more robust: the PDParser (w/o TS) model whose performance decreases rapidly (22.07\%) with the increase of the number of paragraphs, while that of our teacher-student model only decreases by 7.42\% (from 74.51\% to 68.09\%). 

One reason for this significant improvement is the large-scale silver rhetorical structure corpus (MCDTB\_dist) oracle annotated by golden topic structure can better cover complex discourse rhetorical structures. In the MCDTB corpus, there are 27-37 types of discourse rhetorical trees annotated in paragraphs 6-10, which do not increase with the number of paragraphs. However, in MCDTB\_dist corpus, the types of discourse rhetorical structure trees in 6-10 paragraphs have increased from 35 to 437, covering complex discourse rhetorical structure trees better. More details are shown in Figure~\ref{fig: mcdtb_dist vs mcdtb} in Appendix~\ref{sec:appendixc}.

\subsection{Performance on English RST-DT}
We also conduct experiments on RST-DT to demonstrate the generalization of our framework.  Following previous works~\cite{sporleder-lascarides-2004-combining,DBLP:conf/aaai/JiangFCLZ021,huber2022predicting}, we prune and modify the original discourse tree in RST-DT to the macro level to evaluate discourse parsing at the macro level.

Similar to previous work~\cite{huber2022predicting}, we select 5500 documents from the WIKI727K as the topic structure corpus and use the teacher model to generate the silver rhetorical structure corpus (WIKI\_dist) by oracle annotation. We then divide it into 5000 documents as the training set and 500 documents as the validation dataset for the student model. To focus more on the macro-level structure, we use first- and second-level section names as topic boundaries and lower-level section names as paragraph boundaries. 

For the components of the teacher-student model, we use the TM-BERT model~\cite{DBLP:conf/aaai/JiangFCLZ021} as the teacher model and the latest SOTA model~\cite{kobayashi2022simple} as the student model, which is a top-down discourse parser based on DeBERTa~\cite{he2020deberta}~\footnote{All hyper-parameters are default in the published paper.}.

\subsubsection{Baselines}
\textbf{Distantly Supervised Methods.} 

\textbf{Parser$_{senti.}$}~\cite{huber-carenini-2020-MEGA} and \textbf{Parser$_{summ.}$}~\cite{xiao-etal-2021-predicting} are two distantly supervised methods using document-level sentiment information and the attention matrix of the summarization model to build the discourse tree. \textbf{TS$_{wiki}$}~\cite{huber2022predicting} is the SOTA distantly supervised method parsing discourse rhetorical trees by converting results from topic segmentation.

\textbf{Supervised Methods.} 

\textbf{SL04}~\cite{sporleder-lascarides-2004-combining} is the first greedy bottom-up method to build macro-level discourse trees on the RST-DT. \textbf{WL17}~\cite{wang-etal-2017-two} is a discourse parser based on the traditional SVM model and builds the discourse tree with the shift-reduce algorithm. \textbf{PDParser (w/ auto TS)}~\cite{DBLP:conf/aaai/JiangFCLZ021} is a discourse parser using the synthetic topic structure to build the discourse tree. \textbf{SpanBERT}~\cite{guz-carenini-2020-coreference} is one SOTA method based on the pre-trained language model (SpanBERT). It also uses the shift-reduce algorithm to build the discourse tree. \textbf{DeBERTa}~\cite{kobayashi2022simple} is the latest SOTA model, which uses the DeBERTa as the local model to build the discourse tree from top to down.

\subsubsection{Results}
Table~\ref{tab:dist_rstdt} shows the performance of our proposed methods and baselines on English RST-DT. Similar to that in Chinese MCDTB, our proposed teacher-student model achieves the best performance (44.42\%) among a variety of distant supervised models. Moreover, by unifying the in- and out-of-domain annotation form and learning objective, the teacher-student model with oracle annotation also achieves the best performance among supervised models, especially 3.37\% higher than the latest SOTA model (DeBERTa). It demonstrates the effectiveness of our proposed method.

\begin{table}[htbp]
\begin{center}
\resizebox{\linewidth}{!}{
\begin{tabular}{lll}
\hline
\textbf{Scenario}      & \textbf{Method}  & \textbf{Span} \\ \hline
Distantly Supervised & Parser$_{senti.}$          & 31.62         \\
Distantly Supervised & Parser$_{summ.}$          & 32.09         \\
Distantly Supervised & TS$_{Wiki}$          & 41.90         \\
Distantly Supervised & Teacher-Student (ours) & \underline{44.42 }        \\\hline
Supervised           & SL04            & 34.29         \\
Supervised           & WL17            & 37.40         \\
Supervised           & PDParser (w/ auto TS)    & 40.52         \\ 
Supervised           & SpanBERT    & 52.75         \\
Supervised           & DeBERTa    & 54.81        \\

Supervised           & Teacher-Student (ours) & \underline{\textbf{58.18}}             \\ \hline
\end{tabular}
}
\caption{The performance on RST-DT.}
\label{tab:dist_rstdt}
\end{center}
\end{table}

\section{Conclusion}

In this paper, we propose a topic-driven distant supervision framework for macro-level discourse parsing by exploring the relation between topic structure and rhetorical structure\footnote{https://github.com/fjiangAI/TDSF\_DP}. In the framework, we propose two novel methods based on transfer learning and teacher-student models that narrow the distance between in- and out-of-domain tasks from the consistency of the model and the corpus and can fully leverage both in- and out-of-domain samples for retraining. Experimental results show that the proposed method achieves the best performance in both Chinese MCDTB and English RST-DT datasets, demonstrating the effectiveness of our framework. In future research, we will jointly learn the rhetorical and topic structure and analyze the discourse structure of the text more comprehensively. 

\section*{Limitations}

In this paper, we are concerned about the completeness of the silver rhetorical structure corpus we constructed. Despite being annotated with both topic and rhetorical structure, the MCDTB\_dist and WIKI\_dist corpus is not entirely correct, as its rhetorical structure was constructed through oracle annotation. We aim to improve its quality and incorporate human input in future work. Furthermore, we plan to expand its unannotated attributes, such as nuclearity and the rhetorical relationship between discourse units, to better represent the discourse structure of the text.



\bibliography{anthology,custom}

\begin{thebibliography}{31}
\expandafter\ifx\csname natexlab\endcsname\relax\def\natexlab#1{#1}\fi

\bibitem[{Badjatiya et~al.(2018)Badjatiya, Kurisinkel, Gupta, and
  Varma}]{badjatiya2018attention}
Pinkesh Badjatiya, Litton~J Kurisinkel, Manish Gupta, and Vasudeva Varma. 2018.
\newblock Attention-based neural text segmentation.
\newblock In \emph{European Conference on Information Retrieval (ECIR)}, pages
  180--193.

\bibitem[{Carlson et~al.(2003)Carlson, Marcu, and
  Okurowski}]{carlson2003building}
Lynn Carlson, Daniel Marcu, and Mary~Ellen Okurowski. 2003.
\newblock Building a discourse-tagged corpus in the framework of rhetorical
  structure theory.
\newblock \emph{Current and New Directions in Discourse and Dialogue}, pages
  85--112.

\bibitem[{Cohan and Goharian(2018)}]{cohan2018scientific}
Arman Cohan and Nazli Goharian. 2018.
\newblock Scientific document summarization via citation contextualization and
  scientific discourse.
\newblock \emph{International Journal on Digital Libraries}, 19(2):287--303.

\bibitem[{Devlin et~al.(2019)Devlin, Chang, Lee, and
  Toutanova}]{devlin-etal-2019-bert}
Jacob Devlin, Ming-Wei Chang, Kenton Lee, and Kristina Toutanova. 2019.
\newblock \href {https://doi.org/10.18653/v1/N19-1423} {{BERT}: Pre-training of
  deep bidirectional transformers for language understanding}.
\newblock In \emph{Proceedings of the 2019 Conference of the North {A}merican
  Chapter of the Association for Computational Linguistics: Human Language
  Technologies, Volume 1 (Long and Short Papers)}, pages 4171--4186,
  Minneapolis, Minnesota. Association for Computational Linguistics.

\bibitem[{Glavaš and Somasundaran(2020)}]{glava2020twolevel}
Goran Glavaš and Swapna Somasundaran. 2020.
\newblock Two-level transformer and auxiliary coherence modeling for improved
  text segmentation.
\newblock In \emph{Proceedings of the AAAI Conference on Artificial
  Intelligence (AAAI)}, pages 7797--7804.

\bibitem[{Guz and Carenini(2020)}]{guz-carenini-2020-coreference}
Grigorii Guz and Giuseppe Carenini. 2020.
\newblock \href {https://doi.org/10.18653/v1/2020.codi-1.17} {Coreference for
  discourse parsing: A neural approach}.
\newblock In \emph{Proceedings of the First Workshop on Computational
  Approaches to Discourse}, pages 160--167, Online. Association for
  Computational Linguistics.

\bibitem[{He et~al.(2020)He, Liu, Gao, and Chen}]{he2020deberta}
Pengcheng He, Xiaodong Liu, Jianfeng Gao, and Weizhu Chen. 2020.
\newblock Deberta: Decoding-enhanced bert with disentangled attention.
\newblock In \emph{International Conference on Learning Representations}.

\bibitem[{Hearst(1997)}]{hearst-1997-text}
Marti~A. Hearst. 1997.
\newblock \href {https://aclanthology.org/J97-1003} {Text tiling: Segmenting
  text into multi-paragraph subtopic passages}.
\newblock \emph{Computational Linguistics}, 23(1):33--64.

\bibitem[{Huber and Carenini(2019)}]{huber-carenini-2019-predicting}
Patrick Huber and Giuseppe Carenini. 2019.
\newblock \href {https://doi.org/10.18653/v1/D19-1235} {Predicting discourse
  structure using distant supervision from sentiment}.
\newblock In \emph{Proceedings of the 2019 Conference on Empirical Methods in
  Natural Language Processing and the 9th International Joint Conference on
  Natural Language Processing (EMNLP-IJCNLP)}, pages 2306--2316, Hong Kong,
  China. Association for Computational Linguistics.

\bibitem[{Huber and Carenini(2020)}]{huber-carenini-2020-MEGA}
Patrick Huber and Giuseppe Carenini. 2020.
\newblock \href {https://doi.org/10.18653/v1/2020.emnlp-main.603} {{MEGA} {RST}
  discourse treebanks with structure and nuclearity from scalable distant
  sentiment supervision}.
\newblock In \emph{Proceedings of the 2020 Conference on Empirical Methods in
  Natural Language Processing (EMNLP)}, pages 7442--7457, Online. Association
  for Computational Linguistics.

\bibitem[{Huber et~al.(2022)Huber, Xing, and Carenini}]{huber2022predicting}
Patrick Huber, Linzi Xing, and Giuseppe Carenini. 2022.
\newblock Predicting above-sentence discourse structure using distant
  supervision from topic segmentation.
\newblock In \emph{Proceedings of the AAAI Conference on Artificial
  Intelligence (AAAI)}.

\bibitem[{Jiang et~al.(2021)Jiang, Fan, Chu, Li, Zhu, and
  Kong}]{DBLP:conf/aaai/JiangFCLZ021}
Feng Jiang, Yaxin Fan, Xiaomin Chu, Peifeng Li, Qiaoming Zhu, and Fang Kong.
  2021.
\newblock Hierarchical macro discourse parsing based on topic segmentation.
\newblock In \emph{Proceedings of the AAAI Conference on Artificial
  Intelligence (AAAI)}, pages 13152--13160.

\bibitem[{Jiang et~al.(2018)Jiang, Xu, Chu, Li, Zhu, and
  Zhou}]{jiang-etal-2018-MCDTB}
Feng Jiang, Sheng Xu, Xiaomin Chu, Peifeng Li, Qiaoming Zhu, and Guodong Zhou.
  2018.
\newblock \href {https://aclanthology.org/C18-1296} {{MCDTB}: A macro-level
  {C}hinese discourse {T}ree{B}ank}.
\newblock In \emph{Proceedings of the 27th International Conference on
  Computational Linguistics}, pages 3493--3504, Santa Fe, New Mexico, USA.
  Association for Computational Linguistics.

\bibitem[{Kobayashi et~al.(2021)Kobayashi, Hirao, Kamigaito, Okumura, and
  Nagata}]{kobayashi-etal-2021-improving}
Naoki Kobayashi, Tsutomu Hirao, Hidetaka Kamigaito, Manabu Okumura, and Masaaki
  Nagata. 2021.
\newblock \href {https://doi.org/10.18653/v1/2021.naacl-main.127} {Improving
  neural {RST} parsing model with silver agreement subtrees}.
\newblock In \emph{Proceedings of the 2021 Conference of the North American
  Chapter of the Association for Computational Linguistics: Human Language
  Technologies}, pages 1600--1612, Online. Association for Computational
  Linguistics.

\bibitem[{Kobayashi et~al.(2022)Kobayashi, Hirao, Kamigaito, Okumura, and
  Nagata}]{kobayashi2022simple}
Naoki Kobayashi, Tsutomu Hirao, Hidetaka Kamigaito, Manabu Okumura, and Masaaki
  Nagata. 2022.
\newblock A simple and strong baseline for end-to-end neural rst-style
  discourse parsing.
\newblock \emph{arXiv preprint arXiv:2210.08355}.

\bibitem[{Kobayashi et~al.(2019)Kobayashi, Hirao, Nakamura, Kamigaito, Okumura,
  and Nagata}]{kobayashi-etal-2019-split}
Naoki Kobayashi, Tsutomu Hirao, Kengo Nakamura, Hidetaka Kamigaito, Manabu
  Okumura, and Masaaki Nagata. 2019.
\newblock \href {https://doi.org/10.18653/v1/D19-1587} {Split or merge: Which
  is better for unsupervised {RST} parsing?}
\newblock In \emph{Proceedings of the 2019 Conference on Empirical Methods in
  Natural Language Processing and the 9th International Joint Conference on
  Natural Language Processing (EMNLP-IJCNLP)}, pages 5797--5802, Hong Kong,
  China. Association for Computational Linguistics.

\bibitem[{Koshorek et~al.(2018)Koshorek, Cohen, Mor, Rotman, and
  Berant}]{koshorek-etal-2018-text}
Omri Koshorek, Adir Cohen, Noam Mor, Michael Rotman, and Jonathan Berant. 2018.
\newblock \href {https://doi.org/10.18653/v1/N18-2075} {Text segmentation as a
  supervised learning task}.
\newblock In \emph{Proceedings of the 2018 Conference of the North {A}merican
  Chapter of the Association for Computational Linguistics: Human Language
  Technologies, Volume 2 (Short Papers)}, pages 469--473, New Orleans,
  Louisiana. Association for Computational Linguistics.

\bibitem[{Li et~al.(2018)Li, Sun, and Joty}]{li-2018-segbot}
Jing Li, Aixin Sun, and Shafiq Joty. 2018.
\newblock Segbot: A generic neural text segmentation model with pointer
  network.
\newblock In \emph{Proceedings of the 27th International Joint Conference on
  Artificial Intelligence (IJCAI)}, pages 4166--4172.

\bibitem[{Lukasik et~al.(2020)Lukasik, Dadachev, Papineni, and
  Sim{\~o}es}]{lukasik-etal-2020-text}
Michal Lukasik, Boris Dadachev, Kishore Papineni, and Gon{\c{c}}alo Sim{\~o}es.
  2020.
\newblock \href {https://doi.org/10.18653/v1/2020.emnlp-main.380} {Text
  segmentation by cross segment attention}.
\newblock In \emph{Proceedings of the 2020 Conference on Empirical Methods in
  Natural Language Processing (EMNLP)}, pages 4707--4716, Online. Association
  for Computational Linguistics.

\bibitem[{Mann and Thompson(1987)}]{mann1987rhetorical}
William~C Mann and Sandra~A Thompson. 1987.
\newblock \emph{Rhetorical structure theory: A theory of text organization}.
\newblock University of Southern California, Information Sciences Institute.

\bibitem[{Mihaylov and Frank(2019)}]{mihaylov-frank-2019-discourse}
Todor Mihaylov and Anette Frank. 2019.
\newblock \href {https://doi.org/10.18653/v1/D19-1257} {Discourse-aware
  semantic self-attention for narrative reading comprehension}.
\newblock In \emph{Proceedings of the 2019 Conference on Empirical Methods in
  Natural Language Processing and the 9th International Joint Conference on
  Natural Language Processing (EMNLP-IJCNLP)}, pages 2541--2552, Hong Kong,
  China. Association for Computational Linguistics.

\bibitem[{Morey et~al.(2017)Morey, Muller, and Asher}]{morey-etal-2017-much}
Mathieu Morey, Philippe Muller, and Nicholas Asher. 2017.
\newblock \href {https://doi.org/10.18653/v1/D17-1136} {How much progress have
  we made on {RST} discourse parsing? a replication study of recent results on
  the {RST}-{DT}}.
\newblock In \emph{Proceedings of the 2017 Conference on Empirical Methods in
  Natural Language Processing}, pages 1319--1324, Copenhagen, Denmark.
  Association for Computational Linguistics.

\bibitem[{Sporleder and Lascarides(2004)}]{sporleder-lascarides-2004-combining}
Caroline Sporleder and Alex Lascarides. 2004.
\newblock \href {https://aclanthology.org/C04-1007} {Combining hierarchical
  clustering and machine learning to predict high-level discourse structure}.
\newblock In \emph{{COLING} 2004: Proceedings of the 20th International
  Conference on Computational Linguistics}, pages 43--49, Geneva, Switzerland.
  COLING.

\bibitem[{Subba and Di~Eugenio(2009)}]{subba-di-eugenio-2009-effective}
Rajen Subba and Barbara Di~Eugenio. 2009.
\newblock \href {https://aclanthology.org/N09-1064} {An effective discourse
  parser that uses rich linguistic information}.
\newblock In \emph{Proceedings of Human Language Technologies: The 2009 Annual
  Conference of the North {A}merican Chapter of the Association for
  Computational Linguistics}, pages 566--574, Boulder, Colorado. Association
  for Computational Linguistics.

\bibitem[{Tan et~al.(2022)Tan, Zhang, Kong, and Zhou}]{tan2022towards}
Xin Tan, Longyin Zhang, Fang Kong, and Guodong Zhou. 2022.
\newblock Towards discourse-aware document-level neural machine translation.
\newblock In \emph{Proceedings of the Thirty-First International Joint
  Conference on Artificial Intelligence, {IJCAI-22}}, pages 4383--4389.
\newblock Main Track.

\bibitem[{Van~Dijk and Kintsch(1983)}]{van1983strategies}
Teun~A Van~Dijk and Walter Kintsch. 1983.
\newblock \emph{Strategies of discourse comprehension}.
\newblock Acadamic Press.

\bibitem[{Wang et~al.(2017)Wang, Li, and Wang}]{wang-etal-2017-two}
Yizhong Wang, Sujian Li, and Houfeng Wang. 2017.
\newblock \href {https://doi.org/10.18653/v1/P17-2029} {A two-stage parsing
  method for text-level discourse analysis}.
\newblock In \emph{Proceedings of the 55th Annual Meeting of the Association
  for Computational Linguistics (Volume 2: Short Papers)}, pages 184--188,
  Vancouver, Canada. Association for Computational Linguistics.

\bibitem[{Xiao et~al.(2021)Xiao, Huber, and
  Carenini}]{xiao-etal-2021-predicting}
Wen Xiao, Patrick Huber, and Giuseppe Carenini. 2021.
\newblock \href {https://doi.org/10.18653/v1/2021.naacl-main.326} {Predicting
  discourse trees from transformer-based neural summarizers}.
\newblock In \emph{Proceedings of the 2021 Conference of the North American
  Chapter of the Association for Computational Linguistics: Human Language
  Technologies}, pages 4139--4152, Online. Association for Computational
  Linguistics.

\bibitem[{Xing et~al.(2020)Xing, Hackinen, Carenini, and
  Trebbi}]{xing-etal-2020-improving}
Linzi Xing, Brad Hackinen, Giuseppe Carenini, and Francesco Trebbi. 2020.
\newblock \href {https://aclanthology.org/2020.aacl-main.63} {Improving context
  modeling in neural topic segmentation}.
\newblock In \emph{Proceedings of the 1st Conference of the Asia-Pacific
  Chapter of the Association for Computational Linguistics and the 10th
  International Joint Conference on Natural Language Processing}, pages
  626--636, Suzhou, China. Association for Computational Linguistics.

\bibitem[{Yu et~al.(2022)Yu, Zhang, Fu, and Zhang}]{yu-etal-2022-rst}
Nan Yu, Meishan Zhang, Guohong Fu, and Min Zhang. 2022.
\newblock \href {https://doi.org/10.18653/v1/2022.acl-long.294} {{RST}
  discourse parsing with second-stage {EDU}-level pre-training}.
\newblock In \emph{Proceedings of the 60th Annual Meeting of the Association
  for Computational Linguistics (Volume 1: Long Papers)}, pages 4269--4280,
  Dublin, Ireland. Association for Computational Linguistics.

\bibitem[{Zhang et~al.(2021)Zhang, Kong, and
  Zhou}]{zhang-etal-2021-adversarial}
Longyin Zhang, Fang Kong, and Guodong Zhou. 2021.
\newblock \href {https://doi.org/10.18653/v1/2021.acl-long.305} {Adversarial
  learning for discourse rhetorical structure parsing}.
\newblock In \emph{Proceedings of the 59th Annual Meeting of the Association
  for Computational Linguistics and the 11th International Joint Conference on
  Natural Language Processing (Volume 1: Long Papers)}, pages 3946--3957,
  Online. Association for Computational Linguistics.

\end{thebibliography}
\bibliographystyle{acl_natbib}

\appendix

\begin{figure*}[htbp]
	\centering
	\includegraphics[width=16cm]{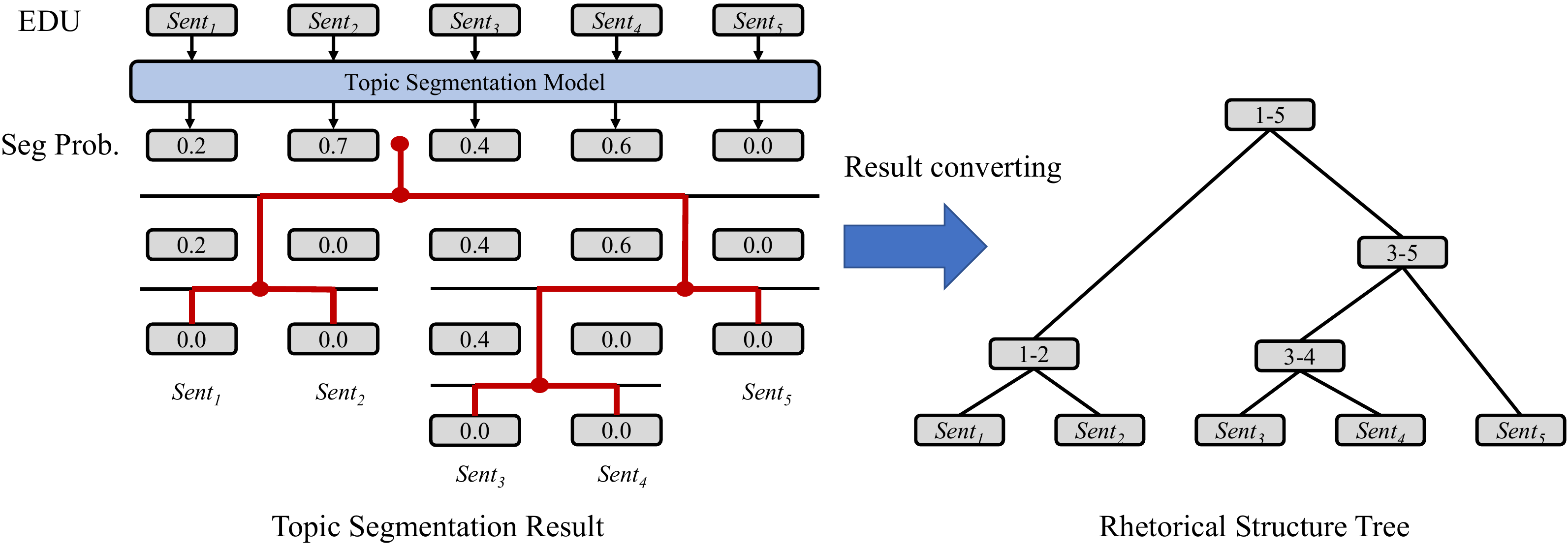}
	\caption{An example of the topic-driven distant supervision by result converting.}\label{fig: topic_segmentation_example}
\end{figure*}

\section{The Process of the Topic-driven Distant Supervision by Result Converting}\label{sec:appendixd}

Figure~\ref{fig: topic_segmentation_example} shows an example of topic-driven distant supervision by result converting. The topic segmentation model could predict the sequence of EDU to get the segmentation probability (Seg Prob.). Then the result converting method will split the sequence according to the order of the probability of segment points. For example, sentence 2 ($Sent_2$) is the highest probability (0.7) that is split first. Then is sentence 4 ($Sent_4$) and sentence 3 ($Sent_3$). Therefore, it uses the top-down parsing method to convert the topic segmentation result into a rhetorical structure tree.

\begin{figure*}[htbp]
	\centering
	\begin{subfigure}[t]{7.5cm}
		\centering
		\includegraphics[width=7.5cm]{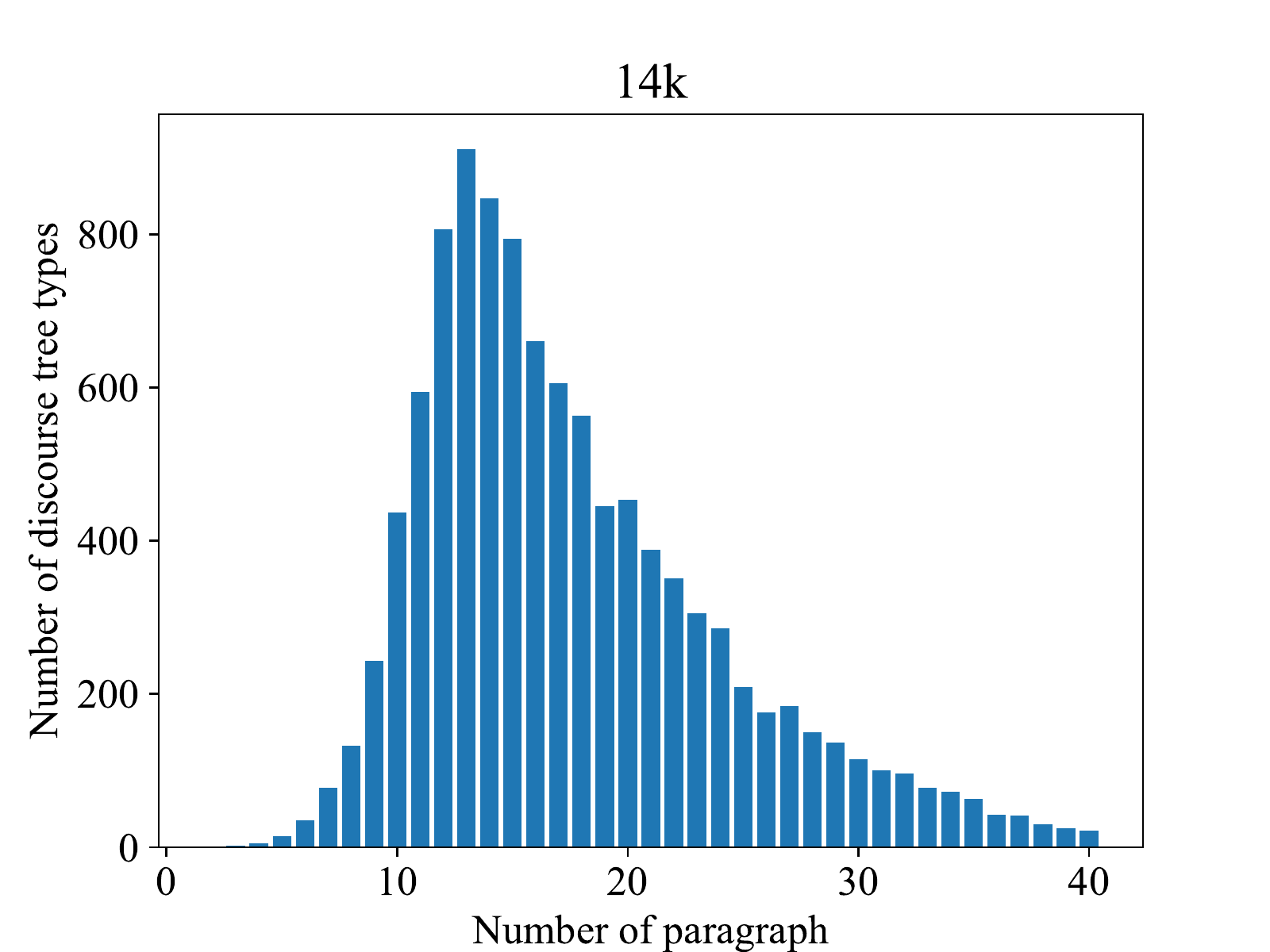}
		\subcaption{ MCDTB\_dist}
	\end{subfigure}
        \quad
	\begin{subfigure}[t]{7.5cm}
		\centering
		\includegraphics[width=7.5cm]{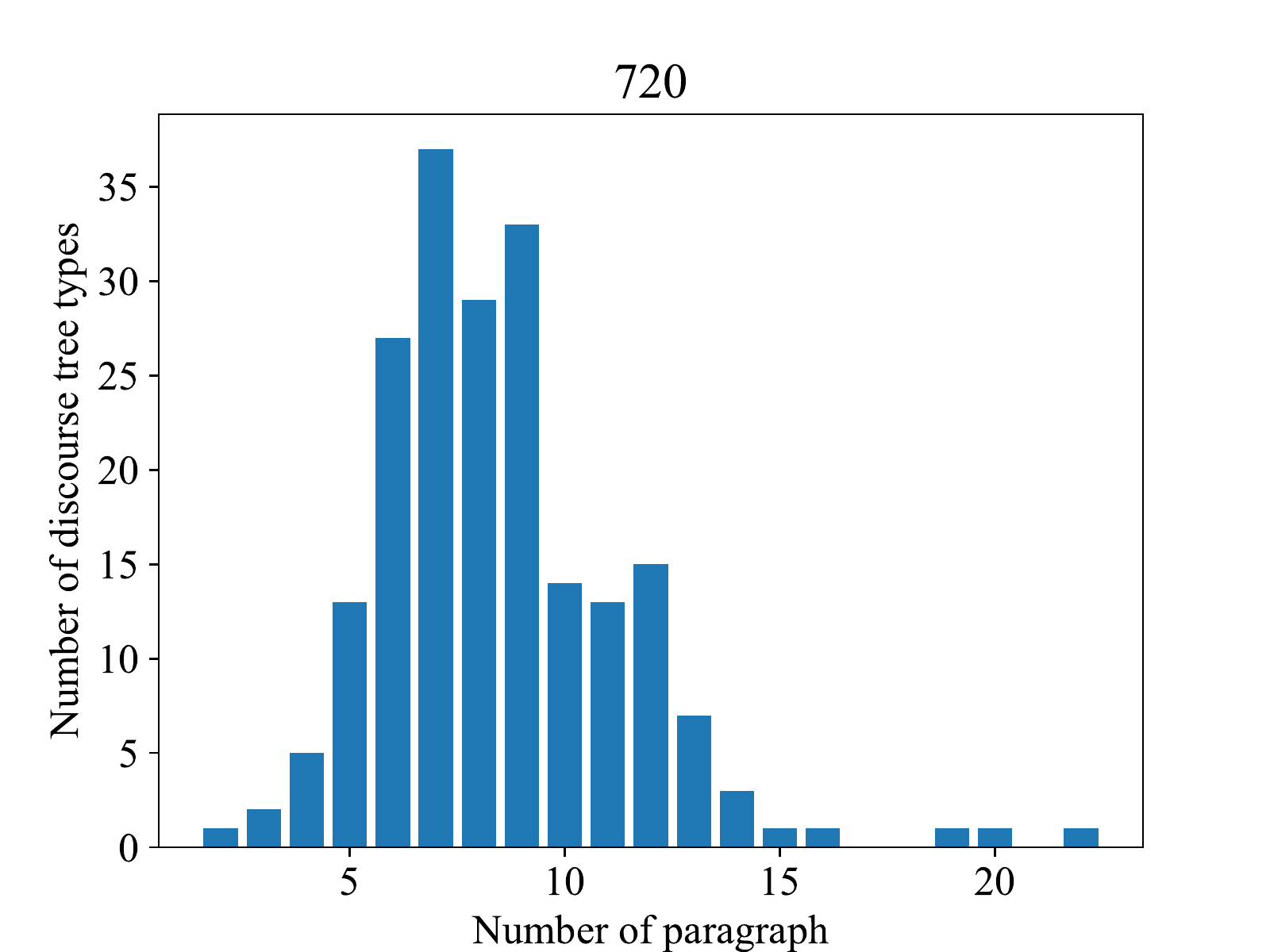}
		\subcaption{MDCTB}
	\end{subfigure}
	\quad
	\caption{The main distribution of discourse tree types in MDCTB\_dist and MCDTB.}\label{fig: mcdtb_dist vs mcdtb}
\end{figure*}

\section{The Details of Silver Rhetorical Structure Corpus}\label{sec:appendixa}

We construct our macro topic structure corpus for the out-of-domain task using the Gigaword~\footnote{https://catalog.ldc.upenn.edu/LDC2009T2} and WIKI727K~\cite{koshorek-etal-2018-text} corpora as our data sources. In Chinese, we select 14,393 news documents with subheadings from the Xinhua News Agency news in the Gigaword corpus and use the subheadings as topic boundaries. In English, we use 5,500 wiki documents with section names, following previous work~\cite{huber2022predicting}, and use the first- and second-level section names as topic boundaries and lower-level section names as paragraph boundaries.

In our transfer learning method, we train an out-of-domain topic segmentation model using the topic structure corpus and then map the labels to convert it into a rhetorical tree construction model. 

In the teacher-student model, we use a ten-fold cross-validation method to oracle annotate the topic structure corpus into a silver rhetorical structure corpus (MCDTB\_dist and WIKI\_dist). It means that we split the dataset into 10 folds, and the silver rhetoric structure on each fold is obtained by the topic segmentation model trained by the remaining nine datasets through the oracle annotation method.

\section{Experimental Settings}\label{sec:appendixb}
\subsection{MCDTB}

The hyper-parameters of the topic segmentation model used are following the previous work~\cite{DBLP:conf/aaai/JiangFCLZ021}: batch-size=2, epoch=10, max-length=512, and learning rate=1e-5. The pre-trained language model is the bert-base model (https://huggingface.co/bert-base-chinese).

In the teacher-student model we proposed, the main hyper-parameters of the student model (BLINK) are the following: the batch-size=2, epoch=50, the hidden size of GRU is 64, the layer number of GRU is 4, and the learning rate=1e-6. The pre-trained language model is the chinese-xlnet-mid model (https://huggingface.co/hfl/chinese-xlnet-mid).

We use an NVIDIA Tesla V100 GPU with 32GB to conduct the experiment.
\subsection{RST-DT}
The hyper-parameter of the topic segmentation model used is the same as the model in MCDTB, except that the pre-trained language model is an English bert-base-uncased model (https://huggingface.co/bert-base-uncased).

In the teacher-student model we proposed, the main hyper-parameters of the student model (Deberta) are the same as previous work~\cite{kobayashi2022simple}.

We use an NVIDIA RTX 3090 GPU with 24GB to conduct the experiment.
 \section{The Main Distribution of Discourse Tree Types in MCDTB\_dist and MCDTB}\label{sec:appendixc}

Figure~\ref{fig: mcdtb_dist vs mcdtb} shows the main distribution of discourse tree types in MCDTB\_dist and MCDTB. In MCDTB\_dist corpus, the discourse tree types increase with the number of paragraphs when the document has less than 13 paragraphs. Utilizing various types of discourse rhetorical structure trees can lead to a more robust structure tree construction model and improved performance. Additionally, even though there may be a decline in diversity in longer documents (\#paragraphs > 13), it is still significantly more than the types in manually annotated MCDTB. For instance, documents with 25 paragraphs still contain over 200 different types of discourse structure trees in MCDTB\_dist, while MCDTB is basically not covered that.

\end{document}